\documentclass[10pt,twocolumn]{article}

\usepackage[T1]{fontenc}
\usepackage[utf8]{inputenc}
\usepackage{lmodern}
\usepackage{microtype}
\usepackage[margin=0.75in]{geometry}
\usepackage{amsmath,amssymb,mathtools}
\usepackage{booktabs}
\usepackage{graphicx}
\usepackage{subcaption}
\usepackage{enumitem}
\usepackage{xcolor}
\usepackage{xspace}
\usepackage{array}
\usepackage{tikz}
\usetikzlibrary{arrows.meta,positioning,fit,calc,backgrounds}
\usepackage[super,sort&compress]{natbib}
\usepackage{url}

\definecolor{linkblue}{RGB}{20,72,140}
\definecolor{factfill}{RGB}{238,244,252}
\definecolor{thoughtfill}{RGB}{244,240,252}
\definecolor{displayfill}{RGB}{240,248,242}
\definecolor{runtimefill}{RGB}{252,244,235}
\usepackage[
  colorlinks=true,
  allcolors=linkblue,
  breaklinks=true,
  bookmarksnumbered=true
]{hyperref}
\usepackage[nameinlink,capitalise,noabbrev]{cleveref}

\newcommand{\cid}{\textsc{CID}\xspace}
\newcommand{\tct}{\textsc{TCT}\xspace}

\DeclareMathOperator{\Project}{Project}

\newif\ifanonymous
\anonymousfalse

\newcommand{\papertitle}{Continuous Interaction Diffusion: A Diffusion-Native Runtime for Asynchronous Tool-Augmented Reasoning}

\title{\papertitle}

\ifanonymous
  \author{Anonymous Authors}
\else
  \author{%
    Yuhang Cao\\
    Nanjing University\\
    \texttt{caoyuhang@smail.nju.edu.cn}
  }
\fi

\date{August 2026}

\hypersetup{%
  pdftitle={\papertitle},
  pdfauthor={\ifanonymous Anonymous Authors\else Yuhang Cao\fi},
  pdfsubject={Diffusion language models, tool-augmented reasoning, and asynchronous runtimes},
  pdfkeywords={diffusion language models, tool use, asynchronous reasoning, latent reasoning, model-runtime co-design}
}

\begin{document}

\maketitle
\begin{abstract}
Large language models increasingly rely on external tools to access up-to-date information, perform computation, and interact with the outside world. For autoregressive models, tool use naturally fits the generation process: the model emits a tool call, waits for the result, and then continues generating. Diffusion language models (dLLMs), however, reason by repeatedly refining many parts of their output in parallel, making this stop-and-resume interaction pattern unnecessarily restrictive. It can force tool decisions before the model's reasoning has stabilized, delay useful observations until a discrete call finishes, and introduce redundant refinement and tool execution, potentially hurting both task accuracy and inference efficiency.

We introduce \emph{Continuous Interaction Diffusion} (CID), a diffusion-native model--runtime architecture that integrates tool interaction into iterative denoising. CID separates a model-read-only fact channel, a thought channel represented by a \emph{Typed Cognitive Tensor}, and a display channel. Information needs can emerge before a textual or JSON call is fully serialized, allowing \emph{perceptual bindings} to launch external reads while denoising continues. Returned results are projected into the evolving thought state and can revise earlier cognition and display regions. Persistent bindings reuse static results without repeated external execution and refresh changing sources when needed. CID is designed to expose evidence earlier, overlap tool latency with model computation, reduce duplicate external work, and preserve useful computation after new evidence arrives. We formalize the architecture, runtime, and training objectives, and define an evaluation protocol for task quality and end-to-end efficiency. This first paper focuses on read-only tools and makes no empirical performance claims.
\end{abstract}

\section{Introduction}
\label{sec:introduction}

Language models increasingly rely on external tools for information and capabilities that are difficult or impossible to obtain from model parameters alone. A model may search the Web for recent facts, read a file, query a database, call a calculator, or inspect the state of a software environment. In current systems, these interactions are usually organized as a sequence of explicit rounds: the model produces some reasoning, emits a structured tool request, waits for the external system to return a result, appends that result to its context, and then resumes generation. ReAct-style agents make this reasoning--acting alternation explicit \citep{yao2022react}, while Toolformer-style training teaches models to insert API calls directly into token sequences \citep{schick2023toolformer}. This interface has become a natural default because it closely matches the generation process of autoregressive language models.

An autoregressive language model generates text from left to right. At each step it predicts the next token conditioned on tokens that have already been produced. Once a token has been emitted, the standard decoding process treats it as fixed and moves forward. Tool use therefore fits naturally into the same timeline: generate a call, receive an observation, place the observation after the call, and continue generating later tokens. The model's internal computation, visible text, and external observations are all ordered along one causal sequence.

Diffusion language models (dLLMs) use a different generation process. Instead of committing to one new token at a time, a dLLM starts from a partially unknown or corrupted sequence and repeatedly refines it through denoising steps. Masked diffusion models can fill multiple unresolved positions in parallel, while decoding variants can additionally re-mask or reopen low-confidence positions \citep{nie2025llada,arriola2025blockdiffusion,ye2025dream,huang2025remedi}. Generation is therefore less naturally described as a single left-to-right timeline. At an intermediate denoising step, many parts of the output may remain uncertain while other parts are already relatively stable, and additional computation can revise unresolved or explicitly reopened content.

This difference becomes important once external tools are introduced. Suppose a search result arrives while a dLLM is still refining its answer. The new evidence may affect a hypothesis formed several denoising steps earlier, change which information is still needed, invalidate part of a draft, and alter several output regions at once. A conventional call--wait--observe interface can still be imposed on a dLLM, but it exposes only a small part of the model's revisable state to the external environment. Current dLLM agents largely retain this interface. DLLM-Searcher, for example, decodes a tool-call region early and continues reasoning while search is in flight, improving overlap while preserving an explicit serialized call boundary \citep{zhao2026dllmsearcher}. More broadly, direct use of current dLLMs in conventional agent workflows shows weaknesses in symbolic precision and temporal feedback handling \citep{lu2026bitterlesson}. These observations motivate a tool interface designed around the dynamics of diffusion generation itself.

We propose \emph{Continuous Interaction Diffusion} (\cid), a diffusion-native architecture in which model cognition, external perception, and user-visible generation evolve concurrently. The central abstraction is a \emph{persistent perceptual binding}. When the model continues to need some information, the runtime maintains a binding between that need and an external source. If the source is static, the runtime can reuse the cached result while repeatedly re-projecting its meaning into the evolving thought state. If the source changes, the runtime can refresh or stream new observations. The model therefore does not have to repeatedly serialize the same request merely to keep the corresponding information cognitively active.

CID separates the system state into three coupled channels. The \emph{fact channel} contains information controlled outside the generative state, such as user-pinned constraints or authoritative external values; the model may read these facts but cannot overwrite them. The \emph{thought channel} contains hypotheses, plans, unresolved information needs, tool intentions, interpretations, uncertainty, and intermediate conclusions. We represent it as a \emph{Typed Cognitive Tensor} (\tct), a continuous locally diffused structure augmented with role distributions, symbolic anchors, source links, uncertainty, and per-cell noise levels. The \emph{display channel} is the evolving token canvas that converges to the user-visible response. These channels can change together while obeying different representations and write permissions.

This paper formalizes the architecture and makes the design precise enough to implement and falsify. Its contributions are:
\begin{itemize}[leftmargin=*]
  \item We formulate the mismatch between turn-based tool protocols and continuously revisable diffusion generation, and define continuous tool interaction as a model--runtime problem.
  \item We introduce a three-channel architecture with an externally controlled fact channel, a typed continuous thought channel, and a discrete display channel.
  \item We define persistent perceptual bindings that distinguish repeated cognitive refresh from repeated external execution, supporting both static and changing information sources.
  \item We specify an asynchronous runtime, local diffusion clocks, model-side intent interface, training objectives, and a concrete evaluation protocol for read-only tools.
\end{itemize}

The remainder of the paper reviews the necessary background and derives the design requirements in \cref{sec:background}, formalizes \cid in \cref{sec:method}, specifies an empirical evaluation protocol in \cref{sec:experiments}, relates the architecture to prior work in \cref{sec:related-work}, and discusses limitations in \cref{sec:discussion}.

\section{Background and Design Requirements}
\label{sec:background}

CID connects ideas from language-model generation, tool-augmented agents, diffusion models, and continuous latent reasoning. This section introduces the parts of those areas needed to understand the design. We focus on how information enters the model over time, which parts of the model state can still be revised, and how external observations are represented.

\subsection{Autoregressive Generation and Turn-Based Tool Use}

Most deployed language models are autoregressive (AR). Given an input context $x$ and an output sequence $y=(y_1,\ldots,y_L)$, an AR model factorizes generation as
\begin{equation}
  p(y\mid x)=\prod_{\ell=1}^{L}p(y_\ell\mid x,y_{<\ell}).
  \label{eq:ar-factorization}
\end{equation}
At decoding step $\ell$, the model predicts the next token from the fixed prefix $y_{<\ell}$. Standard decoding therefore has a simple temporal structure: earlier tokens become context for later tokens, and generation advances from left to right. Although implementations may cache hidden states or speculate over future tokens, the model-facing abstraction remains a growing prefix.

Conventional tool use inherits this structure. Let $q$ be a user request, $r_k$ a textual reasoning segment, $a_k$ a serialized action such as a function call, and $o_k$ the resulting observation. A typical trajectory is
\begin{equation}
  q \rightarrow r_1 \rightarrow a_1 \rightarrow o_1
  \rightarrow r_2 \rightarrow a_2 \rightarrow o_2 \rightarrow y.
  \label{eq:turn-based}
\end{equation}
For example, a model answering a question about a document may first generate a request such as \texttt{search(query=\ldots)}, stop or yield control while the request is executed, receive passages as new context, and then continue its answer. ReAct explicitly alternates reasoning and actions \citep{yao2022react}; Toolformer represents API calls inside token sequences and trains the model to decide when they are useful \citep{schick2023toolformer}.

This call--observe pattern is effective because an external result can simply be appended after everything the AR model has already generated. The model then conditions future tokens on the new observation. The same interface also creates a strong commitment boundary: information that arrives later naturally influences future generation, while revising already committed reasoning or output requires an additional mechanism such as regeneration, editing, or a new conversational turn.

Tool requests are also usually \emph{serialized}. Before execution, the system needs a sufficiently concrete action containing a tool identity and arguments, often expressed as JSON or another schema-constrained structure. An internal sense that ``more evidence about this entity is needed'' does not by itself initiate I/O; it must first become an executable request. This distinction between an emerging information need and a fully specified call is central to CID.

\subsection{Asynchronous Tool Interaction}

A tool call need not block the entire software runtime. Event-driven agents can launch external work asynchronously, continue other computation, and handle results when they arrive \citep{ginart2024asynchronous}. Speculative tool calling can start likely requests before the surrounding generation is complete, and streaming systems can overlap model computation with incoming observations \citep{hooper2026realtime}. These techniques reduce idle wall-clock time.

However, asynchronous execution alone does not change the model's representation of interaction. In a typical AR agent, a completed result still becomes a later event in the token history. The runtime may be concurrent while the model-facing state remains an ordered sequence of linguistic commitments. Likewise, if the model needs the same source throughout a long generation, persistence is usually implemented through repeated calls, cached prompt content, or runtime-specific bookkeeping rather than through an explicit model--source relation.

It is therefore useful to separate two questions. The first is \emph{when external work executes}: synchronously or asynchronously. The second is \emph{how the model represents an ongoing need for external information}. CID primarily changes the second interface and then uses asynchronous execution to support it.

\subsection{Diffusion Language Models}

Diffusion language models (dLLMs) generate with iterative refinement instead of strict next-token prediction. The exact formulation varies across models, but masked diffusion provides a useful concrete example. Starting from a clean token sequence $x_0$, a forward corruption process progressively replaces tokens with a mask symbol or otherwise removes information, producing increasingly corrupted states $x_t$. A learned reverse process receives a corrupted state and predicts cleaner token assignments. At generation time, the model starts from a highly masked sequence and applies multiple denoising updates until a complete sequence is obtained.

This process changes the geometry of generation. At an intermediate step, the sequence can contain a mixture of stable tokens, uncertain tokens, and unresolved masked positions. A denoising update may predict several positions in parallel. Some methods can also re-mask or reopen low-confidence positions, allowing earlier guesses to be reconsidered \citep{huang2025remedi}. LLaDA showed that masked diffusion can scale to instruction-following language modeling \citep{nie2025llada}; Block Diffusion combines blockwise autoregressive structure with diffusion-style generation \citep{arriola2025blockdiffusion}; Dream explores diffusion language modeling with flexible decoding order \citep{ye2025dream}. Planned Diffusion further illustrates how a coarse plan can coexist with parallel refinement of different output spans \citep{israel2025planned}.

The important property for this paper is \emph{revisability}. Consider a partially denoised answer with several already readable claims and several unresolved regions. If new evidence arrives, a diffusion process can use that evidence to change multiple unresolved positions during later denoising steps and, when the decoder supports reopening, revisit previously predicted positions. The computation is therefore naturally described as refinement of a current global state, rather than extension of an immutable prefix.

This does not imply that every dLLM automatically supports arbitrary editing, perfect parallelism, or efficient tool use. Different dLLM architectures impose different masking schedules, block structures, and decoding constraints. CID relies only on the broader capability that some regions of the generative state can remain revisable across iterative updates. The architecture then asks how external information should enter such a state while it is still evolving.

\begin{figure*}[t]
  \centering
  \resizebox{0.96\textwidth}{!}{\begin{tikzpicture}[
  font=\small,
  stage/.style={draw, rounded corners=2pt, minimum height=0.72cm, align=center, inner xsep=6pt, inner ysep=3pt},
  event/.style={circle, draw, minimum size=4.6mm, inner sep=0pt},
  line/.style={-{Latex[length=1.8mm]}, thick},
  async/.style={-{Latex[length=1.8mm]}, thick, dashed},
  note/.style={font=\scriptsize, align=center},
  edgeLabel/.style={font=\scriptsize, fill=white, inner sep=1pt}
]
  % Turn-based AR lane: each observation starts another discrete decision point.
  \node[font=\bfseries, anchor=east] at (-0.35,2.15) {Turn-based AR};
  \node[stage, fill=thoughtfill, minimum width=1.95cm] (arreason) at (1.05,2.15) {reason};
  \node[stage, fill=runtimefill, minimum width=2.05cm] (arcall) at (3.70,2.15) {explicit call};
  \node[stage, minimum width=1.55cm] (arwait) at (6.10,2.15) {wait};
  \node[stage, fill=factfill, minimum width=2.05cm] (arobs) at (8.50,2.15) {observation};
  \node[stage, minimum width=1.55cm] (ardone) at (10.85,2.15) {done?};
  \node[stage, fill=displayfill, minimum width=2.15cm] (arfinal) at (13.35,2.15) {final response};
  \draw[line] (arreason) -- (arcall);
  \draw[line] (arcall) -- (arwait);
  \draw[line] (arwait) -- (arobs);
  \draw[line] (arobs) -- (ardone);
  \draw[line] (ardone) -- node[edgeLabel, above, yshift=1pt] {yes} (arfinal);
  \draw[line, rounded corners=3pt] (ardone.north)
    -- node[edgeLabel, right, pos=0.45] {no} ++(0,0.78cm) -| (arreason.north);

  % CID lane: the dLLM trajectory keeps advancing while the runtime performs I/O.
  \node[font=\bfseries, anchor=east] at (-0.35,-0.10) {CID on a dLLM};
  % Representative updates only: ellipses denote arbitrary-length portions of the trajectory.
  \node[stage, fill=thoughtfill, minimum width=1.25cm] (cidi) at (2.05,-0.10) {$s_i$};
  \node[stage, fill=thoughtfill, minimum width=1.45cm] (cidi1) at (3.95,-0.10) {$s_{i+1}$};
  \node[stage, fill=thoughtfill, minimum width=1.25cm] (cidj) at (7.05,-0.10) {$s_j$};
  \node[stage, fill=thoughtfill, minimum width=1.45cm] (cidj1) at (8.95,-0.10) {$s_{j+1}$};
  \node[stage, minimum width=1.95cm] (cidcheck) at (12.25,-0.10) {converged?};
  \node[stage, fill=displayfill, minimum width=2.15cm] (cidfinal) at (14.85,-0.10) {final response};

  % Draw continuous arrows first, then mask only the small area occupied by each ellipsis.
  \draw[line] (0.24,-0.10) -- (cidi.west);
  \draw[line] (cidi) -- (cidi1);
  \draw[line] (cidi1) -- (cidj);
  \draw[line] (cidj) -- (cidj1);
  \draw[line] (cidj1) -- (cidcheck);
  \node[font=\large, fill=white, inner xsep=2pt, inner ysep=0pt] (pre) at (0.60,-0.10) {$\cdots$};
  \node[font=\large, fill=white, inner xsep=2pt, inner ysep=0pt] (mid) at (5.48,-0.10) {$\cdots$};
  \node[font=\large, fill=white, inner xsep=2pt, inner ysep=0pt] (post) at (10.48,-0.10) {$\cdots$};
  \draw[line] (cidcheck.east) -- (cidfinal.west);
  \node[note] at ($(cidcheck.east)!0.5!(cidfinal.west)+(0,0.14cm)$) {yes};
  \draw[line, rounded corners=3pt] (cidcheck.south)
    -- node[edgeLabel, right, pos=0.45] {no} ++(0,-0.66cm)
    -| (post.south);

  \node[note] at (2.05,0.54) {need emerges};
  \node[note] at (7.05,0.54) {assimilate result};

  \node[stage, fill=runtimefill, minimum width=1.90cm] (bind) at (2.05,-1.52) {bind + launch};
  \node[stage, fill=factfill, minimum width=2.35cm] (job) at (4.95,-1.52) {external read\\in flight};

  \draw[line] (cidi.south) -- (bind.north);
  \node[note, anchor=west] at ($(cidi.south)!0.5!(bind.north)+(0.10cm,0)$) {need};
  \draw[line] (bind.east) -- (job.west);
  \draw[line, rounded corners=2pt] (job.east) -- ++(0.52cm,0)
    -| (cidj.south);
  \node[note, anchor=west] at ($(cidj.south)+(0.10cm,-0.43cm)$) {result};

\end{tikzpicture}}
  \caption{Interaction timelines of turn-based autoregressive tool use and CID.}
  \label{fig:interface-timing}
\end{figure*}
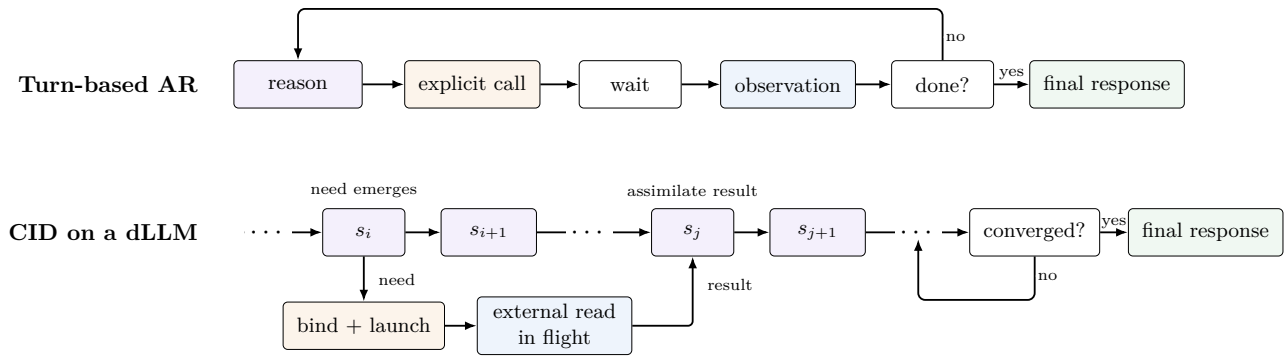

\subsection{Why the Conventional Tool Interface Underuses Diffusion}

Suppose a model is drafting an answer while a search request is running. Before the search result arrives, it may have formed a tentative hypothesis, reserved an output span for evidence, and identified a second question whose relevance depends on the first result. When the result arrives, three kinds of change may be useful simultaneously: revise the hypothesis, update the remaining information needs, and modify an earlier part of the displayed answer.

A turn-based interface represents the same event as a new observation appended after the call. This representation is sufficient for left-to-right continuation, but it does not directly express which earlier cognitive regions the observation should reopen or which ongoing information need the observation satisfies. Current dLLM agents often retain such explicit call regions. DLLM-Searcher, for example, prioritizes decoding a search-call region and lets other reasoning continue while search is in flight \citep{zhao2026dllmsearcher}. This improves overlap, yet the external interaction still begins from a serialized call and returns through a discrete observation boundary.

Existing evaluations also indicate that simply replacing an AR backbone with a dLLM does not automatically produce a strong agent. Current dLLMs can struggle with exact tool schemas and temporally ordered feedback in conventional workflows \citep{lu2026bitterlesson}. A diffusion-native agent interface therefore needs to preserve precise symbolic grounding while giving the runtime access to information needs and revisable state before they become ordinary text.

\Cref{fig:interface-timing} summarizes the mismatch. Beyond asynchronous I/O, CID exposes a need before textual serialization is complete and allows returned evidence to alter multiple still-revisable regions of the same trajectory.

\subsection{Continuous Latent Reasoning}

A related line of work questions whether every intermediate reasoning state must be expressed as natural-language tokens. Textual chains of thought are convenient because they are readable and can be processed by the same language model, but they force intermediate computation through a discrete linguistic representation. Ambiguous hypotheses, multiple possible continuations, or partially formed plans must all be serialized into words before the next reasoning step can use them.

Continuous latent reasoning keeps some intermediate computation in vector-valued model states. Coconut, for example, feeds continuous hidden states back into an autoregressive model as reasoning states and reports that these states can encode multiple possible continuations \citep{hao2024coconut}. LaDiR applies latent diffusion to structured thought representations so that reasoning can be revised more holistically \citep{kang2025ladir}. Such work motivates a continuous thought space in CID.

For tool interaction, however, an unrestricted latent tensor is not enough. The runtime must know whether a latent region represents a hypothesis, an unresolved information need, a source-derived percept, or a stable conclusion. It must also retain exact objects such as file paths, numbers, entity identifiers, source versions, and tool-schema fields. CID therefore combines continuous semantic content with typed roles, symbolic anchors, source links, uncertainty, and local diffusion state.

\begin{table*}[t]
  \centering
  \footnotesize
  \renewcommand{\arraystretch}{0.92}
  \caption{How CID realizes the six design requirements in Section~2.6.}
  \label{tab:cid-mechanisms}
  \begin{tabular}{@{}>{\raggedright\arraybackslash}p{0.22\textwidth}>{\raggedright\arraybackslash}p{0.23\textwidth}>{\raggedright\arraybackslash}p{0.50\textwidth}@{}}
    \toprule
    Design Requirement & CID Mechanism & Effect \\
    \midrule
    Externally Controlled Facts & Three Coupled Channels & Externally controlled values remain outside the model's write authority while thought and display remain revisable. \\
    Precise Symbolic Grounding & Typed Cognitive Tensor & Symbolic anchors and source links keep exact external objects connected to continuous representations. \\
    Early, Non-Linguistic Intent & Latent Information Needs & An information need becomes usable before a complete tool name, query string, or JSON object has converged. \\
    Persistent Perception & Persistent Perceptual Bindings & The same external value can remain active without repeated I/O; changing sources can be refreshed or streamed. \\
    Joint Revision & Perceptual Assimilation & Conflicting evidence can reopen relevant regions while unrelated, well-supported content remains stable. \\
    Continuous Evolution & Asynchronous Runtime & Reasoning and display generation continue while read-only external operations are in flight. \\
    \bottomrule
  \end{tabular}
\end{table*}

\begin{figure*}[t!]
  \centering
  \begin{tikzpicture}[
  font=\small,
  box/.style={draw, rounded corners=2pt, minimum height=1.0cm, align=center, inner sep=6pt},
  arrow/.style={-{Latex[length=2mm]}, thick},
  dashedarrow/.style={-{Latex[length=2mm]}, thick, dashed},
  edgeLabel/.style={font=\scriptsize, inner sep=0pt}
]
  \node[box, fill=factfill, minimum width=4.0cm] (fact) at (-5.6,0)
    {Fact channel $F$\\externally updated, model-read-only};
  \node[box, fill=thoughtfill, minimum width=4.6cm] (thought) at (0,0)
    {Thought channel $T$\\Typed Cognitive Tensor};
  \node[box, fill=displayfill, minimum width=4.0cm] (display) at (5.6,0)
    {Display channel $Y$\\user-visible token canvas};

  \node[box, minimum width=4.0cm] (tools) at (-5.6,-3.6)
    {Read-only sources\\search, files, state, calculators};
  \node[box, fill=runtimefill, minimum width=4.6cm] (runtime) at (0,-3.6)
    {CID runtime\\intent readout, binding lifecycle, refresh};

  \draw[arrow] (fact.east) -- (thought.west);
  \node[edgeLabel] at ($(fact.east)!0.5!(thought.west)+(0,0.78cm)$)
    {protected conditioning};

  \draw[arrow] (thought.east) -- (display.west);
  \node[edgeLabel] at ($(thought.east)!0.5!(display.west)+(0,0.78cm)$)
    {semantic realization};

  \draw[arrow] (display.north) -- ++(0,1.38cm) -| (thought.north);
  \node[edgeLabel] at ($(thought.north)!0.5!(display.north)+(0,1.68cm)$)
    {format/length feedback};

  \draw[arrow] (thought.south) -- (runtime.north);
  \node[edgeLabel, anchor=west] at ($(thought.south)!0.5!(runtime.north)+(0.30cm,0)$)
    {latent needs};

  \draw[arrow] (runtime.west) -- (tools.east);
  \node[edgeLabel] at ($(runtime.west)!0.5!(tools.east)+(0,0.72cm)$)
    {bind / refresh};

  \draw[arrow] (tools.north) -- (fact.south);
  \node[edgeLabel, anchor=east] at ($(tools.north)!0.5!(fact.south)+(-0.30cm,0)$)
    {external updates};

  % Make the read-only-source origin explicit before curving toward thought.
  \draw[dashedarrow] ($(tools.north)+(1.05cm,0)$)
    -- ++(0,0.42cm)
    .. controls (-4.30,-2.15) and (-3.20,-1.22) .. (thought.south west);
  \node[edgeLabel, anchor=west] at (-3.20,-1.62)
    {perceptual projections};
\end{tikzpicture}
  \caption{Overview of the \cid model--runtime architecture.}
  \label{fig:cid-overview}
\end{figure*}
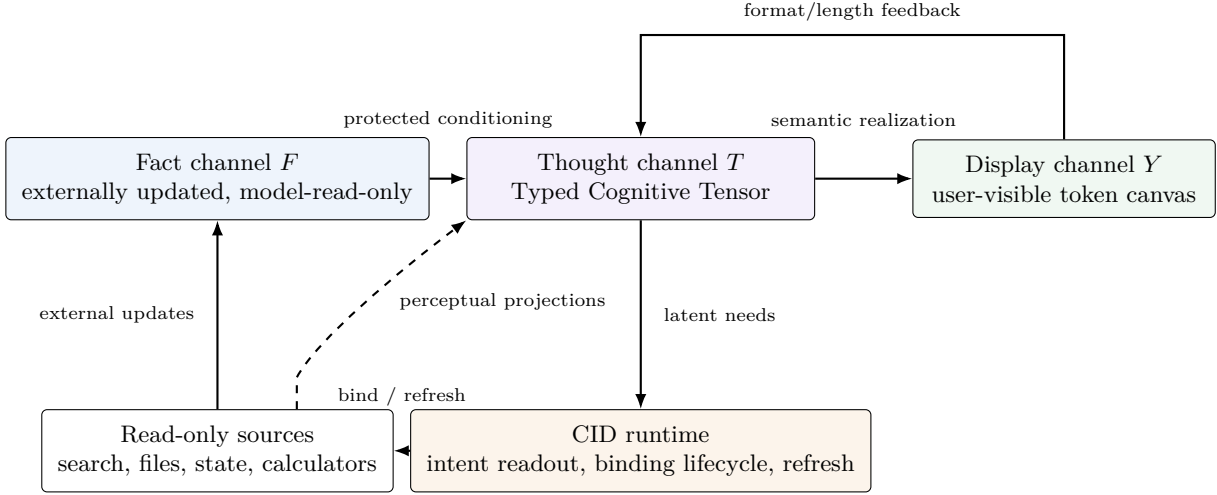

\subsection{Design Requirements}

The preceding background yields six requirements for a diffusion-native interaction architecture.

\paragraph{Externally Controlled Facts.}
Some values must remain outside the model's write authority, including user-pinned constraints, exact source values, and runtime-maintained state, while thought and display remain revisable. Such values can still change when their external source changes. Read-only therefore describes the model's permission over the value, not whether the value is constant over time.

\paragraph{Precise Symbolic Grounding.}
Continuous representations must remain connected to exact external objects. Paths, numbers, entity identifiers, tool schemas, source versions, and provenance should be preserved through symbolic anchors and explicit source links instead of relying on latent similarity alone.

\paragraph{Early, Non-Linguistic Intent.}
A coarse information need should become usable before a complete tool name, query string, or JSON object has converged. Tool identity has been shown to be linearly readable from internal activations in autoregressive models \citep{wu2026toolreadable}, providing evidence that at least source or tool choice can become readable before serialized emission. CID treats the richer typed need interface, including partial arguments and dependencies, as a capability to train and evaluate rather than an established consequence of that result.

\paragraph{Persistent Perception.}
An information need may remain relevant across many model updates. The system should represent this persistence explicitly. For a static source, the same external value can remain active without repeated I/O. For a changing source, the binding can request refreshed values or consume a stream.

\paragraph{Joint Revision.}
New observations should be able to affect several thought and display regions in one evolving state. The system needs selective reopening so conflicting evidence can make relevant regions revisable while unrelated, well-supported content remains stable.

\paragraph{Continuous Evolution.}
Reasoning and display generation should continue while read-only external operations are in flight. When a result arrives, it should condition the current trajectory directly, without requiring the system to restart reasoning as a separate model turn.

\section{CID Design}
\label{sec:method}

CID is designed around two practical goals: improve task quality by allowing external evidence to correct uncertain or stale model state as soon as it becomes available, and improve end-to-end efficiency by overlapping external work with useful denoising while avoiding duplicate tool execution and unnecessary recomputation. The architecture follows from three corresponding requirements. First, exact externally supplied information must remain stable even while model cognition is revised. Second, an information need should become actionable before the model has fully serialized a textual or JSON tool call, so external latency can start earlier. Third, once a result has arrived, the model should be able to keep using or reinterpret it without repeatedly executing the same external operation. The components below implement these requirements jointly rather than treating tool use as a separate pause between generation rounds. Table~\ref{tab:cid-mechanisms} maps the six design requirements in Section~2.6 to the corresponding CID mechanisms and summarizes their intended effects.

Given a task prompt $P$, at diffusion update $s$, \cid maintains
\begin{equation}
  \mathcal{S}_s = (F_s, T_s, Y_s, \mathcal{B}_s, \mathcal{J}_s),
  \label{eq:cid-state}
\end{equation}
where $F_s$ is the fact channel, $T_s$ the thought channel, $Y_s$ the display channel, $\mathcal{B}_s$ the active perceptual bindings, and $\mathcal{J}_s$ the in-flight external jobs. The original prompt $P$ is fixed token-level conditioning across the trajectory rather than a fourth mutable channel; explicitly protected prompt content may additionally be represented in $F_s$. The model updates $T_s$ and $Y_s$, while the environment and runtime update $F_s$, $\mathcal{B}_s$, and $\mathcal{J}_s$ under explicit permissions.

This separation makes ownership explicit. Externally controlled values need not be regenerated and therefore cannot silently drift during denoising, while cognition and display remain revisable. Keeping bindings and in-flight jobs in the same system state also lets external work advance without stopping the model trajectory.

\paragraph{Running Example.}

Consider a user who asks the model to compare two recently released systems and cite exact latency numbers from their documentation. At the start of generation, the model can already organize parts of the answer that do not depend on retrieval: it may identify the comparison dimensions, reserve display regions for evidence, and keep the numerical claims unresolved. During the same denoising process, a region of $T_s$ can begin to represent a need for the latency specification of the first system. This need can become useful to the runtime before the model has produced a complete textual request such as a search query or JSON call.

Once the intent adapter assigns sufficient probability to a registered documentation source and enough arguments are bound, the runtime creates a perceptual binding and launches the read asynchronously. Model denoising continues while the read is in flight. The thought state can refine the comparison plan, and the display state can develop stable material whose correctness does not depend on the missing number. Cells that depend on the unavailable evidence remain uncertain rather than being forced to guess a value merely because generation has advanced.

Suppose the documentation then returns an exact latency of 37~ms. The runtime stores the returned value together with its provenance and version information. Policy may promote the exact value to the fact channel when it must be preserved verbatim. The percept encoder simultaneously constructs a projection conditioned on the current thought and display states. A numerical claim cell can become grounded in 37~ms, a tentative hypothesis that assumed a lower latency can reopen, and an already drafted comparison sentence can be revised during subsequent display denoising. These changes can occur within the same evolving trajectory because the observation conditions the current state rather than starting a separate reasoning turn.

If the answer continues to rely on the same specification, the binding stays active. Later denoising updates can re-project the cached 37~ms value without reading the documentation again, preserving its cognitive influence while the surrounding argument changes. If the source is versioned and a newer document appears, the same binding can request an external refresh and assimilate the new value. When the information need becomes inactive and its dependent output is stable, the runtime retires the binding. This example separates four events that conventional tool interfaces often collapse into one textual call: the emergence of a need, binding to a source, arrival of an external value, and continued cognitive use of that value.

\subsection{Three Coupled Channels}
\label{sec:channels}

\paragraph{Fact Channel.}
$F_s$ contains information whose value is controlled outside the generative state. The model can attend to it but cannot overwrite it. Importantly, this does not imply temporal immutability:
\begin{equation}
  F_{s+1} = \operatorname{ExternalUpdate}(F_s, e_{s+1}).
  \label{eq:fact-update}
\end{equation}
A clock, system load, changing document, or user-updated constraint may therefore change between diffusion steps. The channel may contain user-pinned statements, exact values that must be quoted, authoritative tool results, and provenance metadata. Not every tool result must be promoted to $F_s$; ephemeral or low-value observations may remain in the runtime cache and enter cognition only through a temporary projection.

This channel provides grounding while leaving reasoning to the mutable cognitive state. Separating protected values from mutable cognition prevents exact numbers, constraints, and provenance from being altered by later denoising, while external updates can still replace them when the underlying source genuinely changes. This directly targets factual and copying errors without freezing the rest of the model state.

Each fact item is represented as
\begin{equation}
  f_j = (v_j, \kappa_j, t_j, \nu_j, p_j),
\end{equation}
where $v_j$ is the value, $\kappa_j$ its source type, $t_j$ a timestamp, $\nu_j$ a version or freshness marker, and $p_j$ provenance. Fact-channel policy is an explicit runtime decision governing whether a tool result is promoted.

\paragraph{Thought Channel.}
The thought channel is a structured cognitive field $T_s=(c_{s,1},\ldots,c_{s,N})$, with semantic matrix $H_s\in\mathbb{R}^{N\times d}$ whose row $i$ is $h_{s,i}$. It contains reasoning, hypotheses, information needs, tool intentions, percept interpretations, plans, constraints, and conclusions. Unlike text scratchpads or untyped hidden caches, it exposes structured mutable state through a stable model--runtime contract.

Making these states explicit gives the runtime something more useful than a partially generated string to act on. In particular, it can detect an information need, track which regions depend on it, and preserve uncertainty while a result is still in flight. This enables earlier tool execution and more selective revision after the result arrives.

\paragraph{Display Channel.}
$Y_s$ is a length-$L$ sequence over $\mathcal{V}\cup\{\textsc{mask}\}$ that converges to the user-visible output. It may be revised as $T_s$ changes. Display properties such as realized output length (e.g., the current EOS position or populated span), remaining budget, formatting constraints, and citation coverage can be fed back into the thought channel, allowing the model to continuously adjust content rather than enforce such constraints only after generation.

Separating display from thought lets CID revise user-visible text when new evidence arrives without requiring all intermediate cognition to be serialized into that text. It also allows already useful parts of the answer to continue converging while evidence-dependent regions remain uncertain, preserving useful computation during tool latency.

The model performs coupled updates
\begin{align}
  T_{s+1} &\sim D_T(T_s \mid P, F_s, Y_s, \mathcal{P}_s, \tau^T_s), \\
  Y_{s+1} &\sim D_Y(Y_s \mid P, F_s, T_{s+1}, \tau^Y_s),
  \label{eq:coupled-denoise}
\end{align}
where $\mathcal{P}_s$ denotes perceptual projections from active bindings and $\tau^T_s,\tau^Y_s$ are local editability schedules for thought cells and display positions, respectively. For $Y_s$, increased editability can be realized by remasking or bounded revision of already visible tokens.

\subsection{Typed Cognitive Tensor}
\label{sec:tct}

The thought channel is a \emph{Typed Cognitive Tensor} (\tct). Cognitive cell $i$ is
\begin{equation}
  c_{s,i} = (h_{s,i}, r_{s,i}, a_{s,i}, q_{s,i}, u_{s,i}, \tau_{s,i}, \ell_{s,i}),
  \label{eq:cognitive-cell}
\end{equation}
with the following fields:
\begin{itemize}[leftmargin=*]
  \item $h_{s,i}\in\mathbb{R}^d$: continuous semantic content;
  \item $r_{s,i}$: a soft role distribution, such as hypothesis, information need, percept, plan, constraint, or conclusion;
  \item $a_{s,i}$: sparse symbolic anchors, including tool identifiers, entities, paths, numeric values, schema fields, or output spans;
  \item $q_{s,i}$: links to fact items, bindings, or other cognitive cells;
  \item $u_{s,i}$: epistemic or binding uncertainty;
  \item $\tau_{s,i}$: a local diffusion level controlling editability;
  \item $\ell_{s,i}$: lifecycle state, such as active, waiting, stable, or retired.
\end{itemize}

\begin{figure}[htbp]
  \centering
  \resizebox{0.98\columnwidth}{!}{\begin{tikzpicture}[
  font=\scriptsize,
  field/.style={draw, rounded corners=2pt, align=center, minimum width=1.72cm, minimum height=0.62cm, inner sep=3pt},
  link/.style={-{Latex[length=1.5mm]}, semithick}
]
  \node[circle, draw, fill=thoughtfill, minimum size=1.75cm, align=center, font=\small\bfseries] (core) at (0,0)
    {cell $c_{s,i}$};

  \node[field, fill=thoughtfill] (sem) at (0,2.0) {semantic content\\$h_{s,i}\in\mathbb{R}^d$};
  \node[field, fill=thoughtfill] (role) at (2.45,1.18) {soft role\\$r_{s,i}$};
  \node[field, fill=factfill] (anchor) at (2.55,-1.08) {symbolic anchors\\$a_{s,i}$};
  \node[field, fill=factfill] (links) at (0,-2.0) {source / cell links\\$q_{s,i}$};
  \node[field, fill=runtimefill] (unc) at (-2.55,-1.08) {uncertainty\\$u_{s,i}$};
  \node[field, fill=runtimefill] (noise) at (-2.45,1.18) {local diffusion\\$\tau_{s,i}$};
  \node[field, fill=displayfill, minimum width=1.58cm] (life) at (3.95,0.0) {lifecycle\\$\ell_{s,i}$};

  \draw[link] (sem) -- (core);
  \draw[link] (role) -- (core);
  \draw[link] (anchor) -- (core);
  \draw[link] (links) -- (core);
  \draw[link] (unc) -- (core);
  \draw[link] (noise) -- (core);
  \draw[link] (life) -- (core);
\end{tikzpicture}}
  \caption{Structure of a Typed Cognitive Tensor cell.}
  \label{fig:tct-anatomy}
\end{figure}
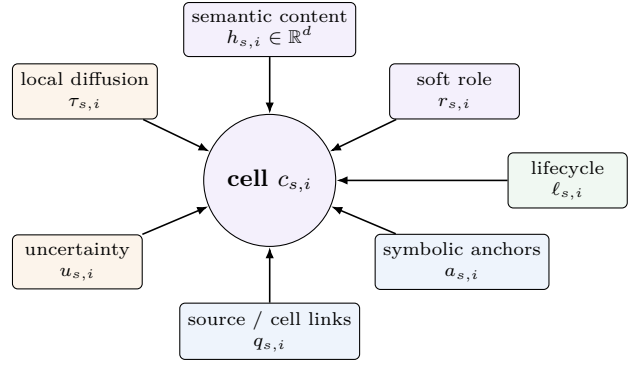

Roles are soft rather than mutually exclusive. A cell can simultaneously represent a hypothesis and the information need that would verify it. Sparse anchors prevent precise objects from drifting in a continuous latent space, while the continuous component preserves ambiguity and multiple competing interpretations.

These fields are included because different parts of tool-augmented reasoning require different behavior. Soft roles retain the flexibility of continuous cognition; symbolic anchors preserve exact tool identifiers, entities, paths, numbers, and output locations; source links identify which state should be revised when evidence changes; uncertainty exposes whether an external result is still needed; and per-cell diffusion levels allow revision to concentrate on affected regions. The expected benefit is higher grounding accuracy with less destructive or global recomputation after each external event.

The local noise vector
\begin{equation}
  \boldsymbol{\tau}_s = (\tau_{s,1},\ldots,\tau_{s,N})
\end{equation}
allows different regions to evolve asynchronously. A supported conclusion may be nearly stable, an unresolved query may remain highly diffused, and a previously stable hypothesis may be reopened after a changing external fact arrives.

\subsection{Latent Information Needs}
\label{sec:latent-needs}

Tool use is represented as a role within $T_s$, not as a separate text region. A model-side intent adapter reads the cognitive field and registered source descriptors $\mathcal{D}=\{d_m\}$:
\begin{equation}
  I_s = G_{\phi}(T_s, Y_s, \mathcal{D}).
  \label{eq:intent-adapter}
\end{equation}
Each candidate need $i_k\in I_s$ contains
\begin{equation}
  i_k = (n_k, \pi_k, \alpha_k, \omega_k, \eta_k, \chi_k),
\end{equation}
where $n_k$ is a continuous need embedding, $\pi_k$ a distribution over registered sources or tools, $\alpha_k$ partially bound arguments, $\omega_k$ uncertainty, $\eta_k$ freshness or persistence demand, and $\chi_k$ links to affected cognitive or display regions.

The adapter exposes a typed interface over the cognitive field. This choice preserves early access to latent intent while insulating the runtime from model-specific hidden-state geometry. It also separates three stages that need not occur at the same diffusion step:
\begin{equation}
  \begin{aligned}
    \text{need emergence} &\rightarrow \text{source selection}\\
                          &\rightarrow \text{argument binding}.
  \end{aligned}
\end{equation}
For read-only tools, a binding may begin before every argument has reached textual certainty, provided the source accepts approximate or progressively refined selectors.

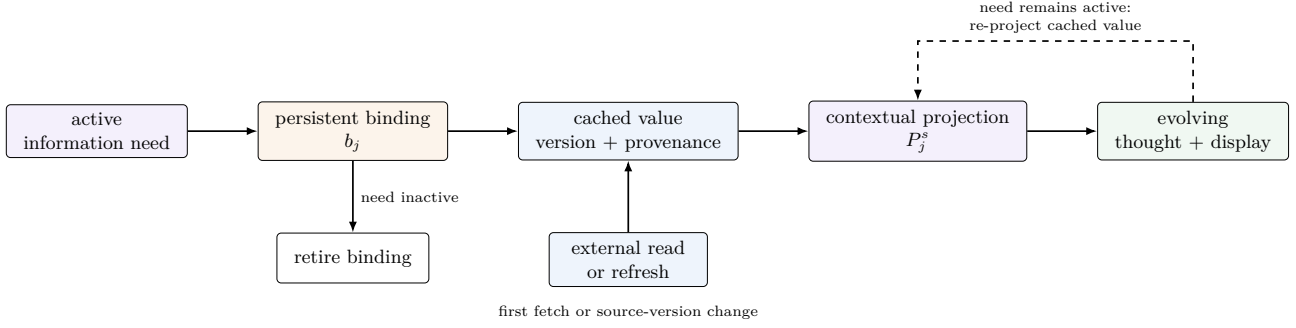
\begin{figure*}[t]
  \centering
  \resizebox{0.96\textwidth}{!}{\begin{tikzpicture}[
  font=\small,
  state/.style={draw, rounded corners=2pt, minimum height=0.82cm, align=center, inner xsep=8pt, inner ysep=4pt},
  flow/.style={-{Latex[length=1.8mm]}, thick},
  loop/.style={-{Latex[length=1.8mm]}, thick, dashed},
  note/.style={font=\scriptsize, align=center}
]
  % Position each state from the previous node's actual border so all four
  % horizontal arrows have the same visible length even when text widens a box.
  \node[state, fill=thoughtfill, minimum width=2.25cm] (need) at (0,0) {active\\information need};
  \node[state, fill=runtimefill, minimum width=2.35cm, right=1.15cm of need] (bind) {persistent binding\\$b_j$};
  \node[state, fill=factfill, minimum width=2.5cm, right=1.15cm of bind] (cache) {cached value\\version + provenance};
  \node[state, fill=thoughtfill, minimum width=2.45cm, right=1.15cm of cache] (project) {contextual projection\\$P_j^s$};
  \node[state, fill=displayfill, minimum width=2.35cm, right=1.15cm of project] (state) {evolving\\thought + display};

  \draw[flow] (need) -- (bind);
  \draw[flow] (bind) -- (cache);
  \draw[flow] (cache) -- (project);
  \draw[flow] (project) -- (state);

  \node[state, minimum width=2.05cm, below=1.18cm of bind] (retire) {retire binding};
  \draw[flow] (bind.south) -- node[note, right] {need inactive} (retire.north);

  \node[state, fill=factfill, minimum width=2.6cm, below=1.18cm of cache] (source) {external read\\or refresh};
  \draw[flow] (source.north) -- (cache.south);
  \node[note] at ($(source.south)+(0,-0.42)$) {first fetch or source-version change};

  \draw[loop] (state.north) -- ++(0,1.0) -| (project.north);
  \node[note] at ($(project.north)!0.5!(state.north)+(0,1.38)$) {need remains active:\\re-project cached value};
\end{tikzpicture}}
  \caption{Lifecycle of a persistent perceptual binding.}
  \label{fig:binding-lifecycle}
\end{figure*}

The main efficiency benefit is that external work can start at need emergence or partial argument binding rather than after a complete call has been decoded. If the need is identified several denoising steps earlier than an executable textual call, those steps can overlap tool latency instead of adding to it. At the same time, retaining uncertainty and partial arguments avoids forcing the model to commit prematurely to details that later denoising may change.

\subsection{Persistent Perceptual Bindings}
\label{sec:bindings}

A conventional tool call is a transient request--response pair. CID instead creates a persistent perceptual binding
\begin{equation}
  b_j = (i_j, m_j, \alpha_j, \rho_j, z_j, \chi_j),
  \label{eq:binding}
\end{equation}
where $i_j$ is the originating need, $m_j$ the source or tool, $\alpha_j$ current arguments, $\rho_j$ a refresh policy, $z_j$ runtime cache and version state, and $\chi_j$ the target cognitive regions.

A binding remains active while the corresponding information need remains active. Repeated intent is therefore not automatically treated as a redundant model failure. It can mean that the model wants the percept to remain cognitively salient. The runtime distinguishes two operations:
\begin{align}
  \text{external refresh: }& v_j^{s+1} \leftarrow m_j(\alpha_j, s+1), \\
  \text{cognitive refresh: }& P_j^{s+1} \leftarrow \Project(v_j, T_s, Y_s).
  \label{eq:two-refreshes}
\end{align}

\paragraph{Static or Unchanged Source.}
If the source version is unchanged, the runtime can reuse $v_j$ but recompute $P_j^{s+1}$. The same fact may therefore receive a different task-specific interpretation as cognition evolves. This supports copying and sustained reference: the source need not be re-read, yet its influence is renewed at each relevant denoising update.

\paragraph{Changing Source.}
If the source is time-varying or streamable, $\rho_j$ may request polling, version checks, event subscriptions, or incremental chunks. An arriving value updates a fact item if policy marks it as protected, and always permits a new cognitive projection. Examples include current time, evolving realized display length, a modified file, sensor input, or streaming retrieval results.

A simple refresh policy is
\begin{equation}
  \rho_j(s)=
  \begin{cases}
    \textsc{reproject}, & \nu_j(s)=\nu_j(s-1),\\
    \textsc{refetch+reproject}, & \nu_j(s)\neq\nu_j(s-1),\\
    \textsc{sleep}, & \Pr(i_j\text{ active})<\delta.
  \end{cases}
  \label{eq:refresh-policy}
\end{equation}
Here $\nu_j(s)$ denotes a version already observed through a source event or a lightweight freshness probe; obtaining it is itself external work. If no cheap version signal exists, a due refresh must re-read the source rather than assume that its version is known. More advanced policies can trade information gain, latency, and tool cost.

This distinction is what allows persistent information use without persistent I/O. For a static source, CID can repeatedly refresh the result's influence on evolving cognition while paying the external access cost only once. For a changing source, the same binding provides an explicit place to decide when a new fetch is worth its latency or cost. Persistent bindings therefore reduce duplicate calls while still allowing long or changing reasoning trajectories to remain grounded in external information.

\subsection{Perceptual Assimilation}
\label{sec:assimilation}

The percept encoder converts each tool result into a context-dependent projection
\begin{equation}
  P_j^s = E_{\psi}(v_j, p_j, T_s, Y_s),
  \label{eq:percept-projection}
\end{equation}
whose interpretation depends on the current thought and display states. Projection cells carry source links to $v_j$ when provenance matters. The thought denoiser integrates them with gated cross-attention or residual updates:
\begin{equation}
  \widetilde{T}_s = T_s + \sum_{j\in\mathcal{B}_s} g_{j,s}\,A(T_s,P_j^s),
  \label{eq:percept-assimilation}
\end{equation}
where $g_{j,s}$ is predicted from need persistence, relevance, and source freshness. This residual form is one realization of the $\mathcal{P}_s$ conditioning in \cref{eq:coupled-denoise}; an implementation may perform the same fusion inside $D_T$.

\begin{figure*}[t]
  \centering
  \resizebox{0.96\textwidth}{!}{\begin{tikzpicture}[
  font=\small,
  model/.style={draw, rounded corners=2pt, fill=thoughtfill, minimum height=0.82cm, align=center, inner xsep=7pt},
  runtime/.style={draw, rounded corners=2pt, fill=runtimefill, minimum height=0.82cm, align=center, inner xsep=7pt},
  source/.style={draw, rounded corners=2pt, fill=factfill, minimum height=0.82cm, align=center, inner xsep=7pt},
  flow/.style={-{Latex[length=1.8mm]}, thick},
  async/.style={-{Latex[length=1.8mm]}, thick, dashed},
  lane/.style={font=\bfseries, anchor=east},
  lab/.style={font=\scriptsize, fill=white, inner sep=1pt}
]
  \node[lane] at (-0.35,1.6) {Model};
  \node[lane] at (-0.35,0.0) {Runtime};
  \node[lane] at (-0.35,-1.6) {Sources};

  \node[model, minimum width=2.45cm] (denoise) at (1.55,1.6) {denoise $T,Y$};
  \node[model, minimum width=2.45cm] (needs) at (5.0,1.6) {expose needs $I$};
  \node[model, minimum width=2.7cm] (assimilate) at (9.1,1.6) {assimilate arrivals};
  \node[model, minimum width=2.65cm] (noise) at (13.1,1.6) {adjust local noise};
  \draw[flow] (denoise) -- (needs);
  \draw[flow] (assimilate) -- (noise);
  \draw[flow, rounded corners=3pt] (noise.north) -- ++(0,0.62) -| (denoise.north);
  \node[lab, anchor=south, yshift=2pt] at (7.3,2.63) {next active update};

  \node[runtime, minimum width=2.55cm, inner xsep=4pt] (match) at (5.0,0) {match / update binding};
  \node[runtime, minimum width=2.8cm, inner xsep=4pt] (schedule) at (9.1,0) {reuse, refresh, or launch};
  \node[runtime, minimum width=2.55cm] (events) at (13.1,0) {consume events};
  \draw[flow] (needs.south) -- (match.north);
  \draw[flow] (match) -- (schedule);
  \draw[flow, rounded corners=3pt] (events.north) -- ++(0,0.38)
    -| ([xshift=0.75cm]assimilate.south);

  \node[source, minimum width=3.2cm] (external) at (9.1,-1.6) {read-only external work};
  \draw[async] (schedule.south) -- node[lab, right] {async} (external.north);
  \draw[async] (external.east) -| (events.south);

  \draw[async] (schedule.north) -- node[lab, anchor=east, xshift=-3pt, align=right] {cached\\re-projection} (assimilate.south);

\end{tikzpicture}}
  \caption{CID runtime loop for read-only perception.}
  \label{fig:cid-runtime}
\end{figure*}
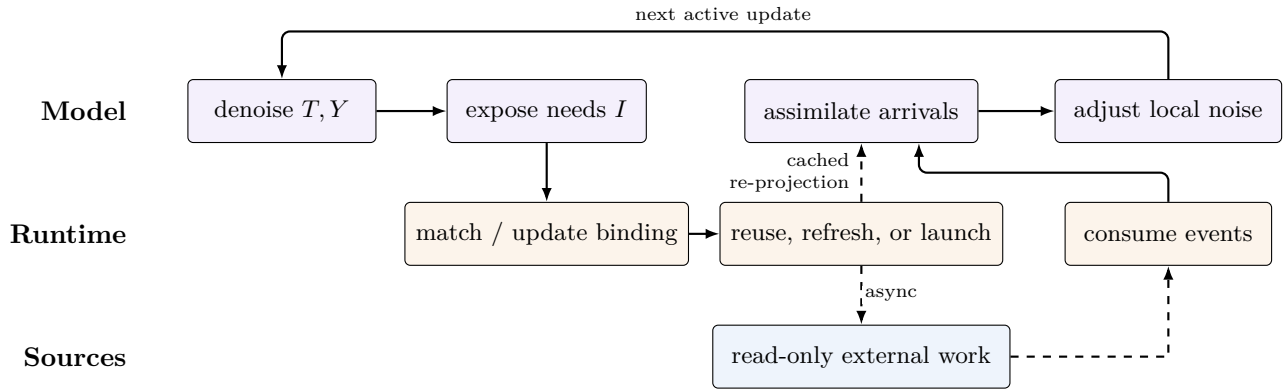

Recomputing the projection from the current $T_s$ and $Y_s$ lets one cached result support different parts of the reasoning as the trajectory evolves, without another external call. It also allows a newly arrived result to affect already formed hypotheses and display regions immediately instead of being useful only for tokens generated afterward.

New evidence may change the editability of existing cells. Let $\Delta_{j,i}$ estimate how strongly percept $j$ changes the posterior of cell $i$. The runtime or model updates
\begin{equation}
  \tau_{s+1,i} = \operatorname{clip}
  \left(\tau_{s,i} + \lambda_{\mathrm{open}}\Delta^-_{j,i}
  - \lambda_{\mathrm{close}}\Delta^+_{j,i}\right),
  \label{eq:local-noise-update}
\end{equation}
where $\Delta^-_{j,i}$ measures conflict and $\Delta^+_{j,i}$ support. Conflicting cognitive cells become more diffused and revisable; supported cells stabilize. Evidence-linked display positions are reopened analogously through $\tau^Y_s$, implemented by remasking or bounded visible-token revision.

Local reopening is important for efficiency as well as correctness. New evidence should revise the state it invalidates, but it should not discard unrelated work that was already correct. Adjusting editability per cell therefore aims to preserve useful computation while concentrating additional denoising on regions that actually need correction.

\subsection{Asynchronous Runtime}
\label{sec:runtime}

\Cref{fig:cid-runtime} describes the runtime loop. Model denoising may continue while read-only work is outstanding, but only until a learned \emph{current-information equilibrium} signal fires. The runtime then pauses if required work remains and resumes on external progress; otherwise it forms a terminal candidate.

The runtime is asynchronous for a direct performance reason: waiting for retrieval, file I/O, or another read-only source should not automatically become idle model time. CID instead allows the model to refine source-independent cognition and display regions while the job is in flight, then revises dependent regions after arrival. End-to-end latency improves when this waiting-time computation remains useful, which is why the evaluation measures latency at matched answer quality rather than token-generation speed alone.

Initialization sets facts $F$, an initial thought state $T$, masked display $Y$, bindings $\mathcal{B}=\varnothing$, and jobs $\mathcal{J}=\varnothing$. Each model update exposes information needs; the runtime matches or updates bindings, launches a read when policy requires a fresh value and no equivalent job is active, consumes events, and re-projects cached values when external access is unnecessary. The model adjusts local noise until current-information equilibrium or a budget boundary; quiescent waiting consumes no model-update budget. Before termination, a freshness barrier completes required bindings and refreshes due values, with new evidence resuming refinement. External work is keyed by source and canonicalized arguments, and a completion applies only when an active binding still has the same work key, making obsolete work cancellable or discardable. Equivalent jobs are deduplicated without suppressing repeated perception: identical needs may share one cached value while receiving fresh projections. A refined need can update a binding's selector, while independent verification can establish another binding. The step cap bounds model updates; a separate wall-clock budget bounds the trajectory.

\medskip
\noindent\rule{\columnwidth}{0.5pt}
\vspace{3pt}
\noindent\textbf{TRAINING OBJECTIVE}\par\vspace{2pt}
CID requires a model trained for channel semantics and asynchronous events. A training trajectory contains a task $x$, protected facts $F$, tool/source descriptors $\mathcal{D}$, external event sequence $E=(e_1,\ldots,e_K)$ with arrival times, target thought structures $T^*$ or weak structural supervision, and final display $Y^*$.

Training randomizes event arrival, source freshness, and cache state. Before a needed event arrives, the model should maintain uncertainty and expose a binding need rather than fabricate a value. Waiting and final snapshots supervise the equilibrium signal; runtime state distinguishes quiescence from termination. After arrival, the model should assimilate the percept, revise affected cells, preserve unrelated cognition, and update the display.

A possible objective is
\begin{align}
  \mathcal{L} ={}& \mathcal{L}_{T}
  + \lambda_Y\mathcal{L}_{Y}
  + \lambda_I\mathcal{L}_{\mathrm{intent}}
  + \lambda_B\mathcal{L}_{\mathrm{bind}} \notag\\
  &+ \lambda_P\mathcal{L}_{\mathrm{assim}}
  + \lambda_R\mathcal{L}_{\mathrm{refresh}}
  + \lambda_G\mathcal{L}_{\mathrm{ground}}
  + \lambda_C\mathcal{L}_{\mathrm{conv}}.
  \label{eq:training-objective}
\end{align}
The first two terms train thought and display diffusion; the others train intent exposure, binding, event-conditioned revision, refresh behavior, symbolic grounding, and the equilibrium signal.

Static-copy training examples repeatedly expose an unchanged source while the output is gradually composed. Dynamic-monitoring examples change the source during denoising and require the thought and display states to track the latest externally supplied value. Search and file-reading examples teach need emergence, partial query binding, delayed arrival, and evidence-conditioned revision.
\par\vspace{3pt}
\noindent\rule{\columnwidth}{0.5pt}
\par

\section{Evaluation Protocol}
\label{sec:experiments}

This paper does not claim empirical performance results. We specify the minimum evidence required to support or reject the CID hypothesis and define how the complete model--runtime system should be evaluated beyond token-generation speed.

\subsection{Research Questions}

\paragraph{RQ1: Do Useful Information Needs Emerge Before Explicit Calls?}
Measure the earliest diffusion step at which the typed intent interface correctly identifies source type and information target. Compare it with the step at which a valid textual or JSON call becomes executable. CID is useful only if latent need detection provides a meaningful lead without excessive false bindings.

\paragraph{RQ2: Can Persistent Bindings Hide External Latency?}
Vary source latency while holding model compute fixed. Use a logical event clock for quality comparisons so accelerator speed cannot change pre-arrival refinement, and report wall-clock latency separately on matched hardware. Measure whether waiting-time computation remains useful after arrival. The central comparison is end-to-end latency at matched answer quality.

\paragraph{RQ3: Does Repeated Perception Improve Assimilation?}
For static sources, compare one-time injection with per-step cognitive re-projection. For dynamic sources, compare no refresh, fixed polling, and need-conditioned refresh. Tasks should test exact copying, evidence application, stale-value avoidance, and revision of already formed output.

\paragraph{RQ4: Is the Typed Cognitive Tensor Necessary?}
Compare (i) a natural-language thought string, (ii) an untyped continuous latent tensor, and (iii) the proposed \tct. Measure accuracy, source-binding stability, symbolic-reference precision, locality of revision, and runtime inspectability.

\paragraph{RQ5: Is Diffusion Essential?}
Compare CID against an asynchronous autoregressive model with the same sources, training data, and approximate compute budget. A convincing result must show a benefit from revising prior cognition or display regions, rather than from asynchronous I/O alone.

\subsection{Task Families}

\paragraph{Static Reference and Copying.}
The model reads a fixed document, table, or code fragment and must repeatedly preserve exact values while composing a longer response. This directly tests the distinction between repeated external execution and repeated cognitive refresh.

\paragraph{Dynamic State Tracking.}
A source changes during generation: current time, a key--value entry, a file version, remaining token budget, or realized display length. The final output must reflect the latest value and revise stale intermediate conclusions.

\paragraph{Delayed Retrieval.}
Question answering tasks provide a retriever whose results arrive after controlled delays. Some answer regions can be prepared without evidence; others must remain unresolved until retrieval arrives.

\paragraph{Streaming Evidence.}
A source emits increasingly informative chunks. The model should revise hypotheses incrementally instead of waiting for the complete result or restarting generation after every chunk.

\paragraph{Competing Sources.}
Multiple bindings expose complementary or conflicting facts. This setting tests source links, uncertainty, repeated verification, and local reopening of cognition.

\subsection{Baselines}

A minimal comparison should include:
\begin{itemize}[leftmargin=*]
  \item autoregressive ReAct with blocking tools;
  \item an event-driven asynchronous autoregressive agent \citep{ginart2024asynchronous,hooper2026realtime};
  \item a dLLM using conventional explicit tool-call regions;
  \item P-ReAct as instantiated by DLLM-Searcher \citep{zhao2026dllmsearcher};
  \item CID with one-time percept injection;
  \item CID with persistent static re-projection;
  \item full CID with static and dynamic refresh;
  \item oracle bindings and arrival-aware local-noise updates as upper bounds.
\end{itemize}

\subsection{Metrics}

\paragraph{Task Quality.}
Use exact match, factual accuracy, evidence support, copying fidelity, and task-specific correctness. For dynamic sources, report stale-value rate.

\paragraph{Interaction Latency.}
Report wall-clock latency, tool-wait overlap, and \emph{intent lead time}
\begin{equation}
  \Delta_{\mathrm{lead}} = t_{\mathrm{explicit\ call}} - t_{\mathrm{binding}}.
\end{equation}

\paragraph{Assimilation.}
Define observation assimilation lag as
\begin{equation}
  \Delta_{\mathrm{assim}} = t_{\mathrm{correct\ state}} - t_{\mathrm{arrival}}.
\end{equation}
Also measure the number and fraction of thought and display regions revised after arrival.

\paragraph{Binding Behavior.}
Report binding precision and recall, persistence duration, external refresh count, cache-hit rate, repeated cognitive projection count, and the fraction of duplicate needs handled without duplicate I/O.

\paragraph{Revision Quality.}
Measure the ratio of corrected errors to newly introduced errors, preservation of unaffected content, and the compute cost of local reopening. These metrics test whether diffusion revision is selective rather than destructive.

\subsection{Key Ablations}

A full evaluation should remove role typing, symbolic anchors, source links, per-cell noise levels, persistent bindings, static re-projection, dynamic source refresh, and display-to-thought feedback. Arrival-time randomization during training should also be ablated to determine whether the system learns genuine asynchronous assimilation or overfits a fixed event schedule.

\subsection{Falsification Criteria}

CID would not be supported if latent bindings do not reliably precede explicit calls, if post-arrival revision destroys more correct state than it repairs, if waiting-time computation is mostly discarded, or if an asynchronous autoregressive baseline matches quality and latency with lower complexity. These outcomes constitute substantive negative results for the CID hypothesis.

\section{Related Work}
\label{sec:related-work}

\paragraph{Diffusion Language Models.}
Early text diffusion work established both continuous and discrete formulations. Diffusion-LM denoises continuous word-vector representations \citep{li2022diffusionlm}, while D3PM develops structured corruption processes for discrete state spaces \citep{austin2021d3pm}. DiffuSeq extends diffusion to conditional sequence-to-sequence text generation \citep{gong2022diffuseq}. Masked diffusion language modeling later showed that a comparatively simple objective can approach autoregressive perplexity \citep{sahoo2024mdlm}. More recent systems scale these ideas toward general-purpose language modeling: LLaDA establishes masked diffusion at large-model scale \citep{nie2025llada}; Block Diffusion interpolates between blockwise autoregression and parallel diffusion \citep{arriola2025blockdiffusion}; and Dream demonstrates flexible decoding order \citep{ye2025dream}. Planned Diffusion partitions output into parallel spans \citep{israel2025planned}, while RemeDi reopens low-confidence tokens during decoding \citep{huang2025remedi}. CID focuses on asynchronous external events as persistent conditions on a jointly revisable cognitive and display state.

\paragraph{Tool-Using Language Models.}
ReAct alternates explicit reasoning and actions \citep{yao2022react}, and Toolformer trains API calls as token sequences \citep{schick2023toolformer}. These methods established tool-augmented reasoning while retaining serialized action boundaries. Asynchronous Tool Usage introduces an event-driven architecture for real-time agents \citep{ginart2024asynchronous}; later work combines asynchronous I/O with speculative calls and streaming-input training \citep{hooper2026realtime}. CID shares the objective of overlapping cognition and I/O, while moving persistent percept state into the model--runtime interface and permitting external events to revise earlier internal and displayed content.

\paragraph{Diffusion Agents.}
DLLM-Searcher trains a dLLM search agent and proposes P-ReAct, which prioritizes decoding a tool-call region so reasoning can continue while search runs \citep{zhao2026dllmsearcher}. This is the closest existing system to CID's latency motivation. CID removes the requirement that every information need first converge into an explicit call region and generalizes a call into a sustained binding with repeated projection or refresh. Evaluations of current dLLMs in agent workflows find weaknesses in strict tool schemas and temporal feedback, while observing stronger performance in non-causal roles such as selection and summarization \citep{lu2026bitterlesson}. CID's typed latent interface and separation between continuous thought and symbolic anchors are intended to address this mismatch, though empirical validation remains necessary.

\paragraph{Latent and Diffusion-Based Reasoning.}
Coconut feeds continuous reasoning states back into an autoregressive model and argues that latent states can represent multiple possible reasoning continuations \citep{hao2024coconut}. Diffusion of Thoughts applies diffusion generation to chains of thought \citep{ye2024dot}. LaDiR constructs structured latent thought blocks and denoises them with latent diffusion \citep{kang2025ladir}. Continuous latent diffusion language models further separate global semantic organization from textual realization \citep{guo2026coladlm}. CID builds on the general case for non-textual intermediate cognition, but introduces typed cells, external source links, persistent perceptual bindings, and a separate display diffusion process.

\paragraph{Internal Tool Intent.}
Recent probing work reports that tool identity can be read and steered from internal activations before a complete call is emitted \citep{wu2026toolreadable}. This supports the feasibility of latent intent readout. CID does not expose arbitrary hidden states directly; it requires a trained, typed adapter whose output is part of the model contract and whose uncertainty can be used by the runtime.

\section{Discussion and Limitations}
\label{sec:discussion}

\paragraph{Architectural Complexity.}
CID adds a continuous thought representation, a typed intent adapter, binding state, event scheduling, and local diffusion control. These components are justified only if they yield benefits unavailable to a simpler asynchronous autoregressive agent. The evaluation in \cref{sec:experiments} therefore treats the AR comparison as a central falsification test.

\paragraph{Supervision for Cognition.}
High-quality targets for typed cognitive cells and binding lifecycles are not naturally available. Possible strategies include distillation from textual reasoning and tool trajectories, weak supervision from source dependencies, synthetic event schedules, and end-to-end learning with auxiliary consistency losses. Each may bias the internal representation toward the teacher or task generator.

\paragraph{Latent-State Stability.}
Typed interfaces make the thought state more controllable than an arbitrary hidden tensor, but roles and anchors may still drift across checkpoints or domains. Dynamic tool registration also requires source descriptors that are expressive enough for generalization yet constrained enough for reliable binding.

\paragraph{Fact-Channel Policy.}
The fact channel is protected from model writes, not guaranteed to be true. A user can pin incorrect information, a trusted source can become stale, and two externally controlled facts can conflict. CID must preserve provenance and uncertainty rather than treating protected information as necessarily correct. Deciding which tool results deserve promotion to the fact channel is itself a policy problem.

\paragraph{Compute and Memory Cost.}
Persistent re-projection may reduce I/O but increase model computation. Local diffusion clocks and cached percept encodings must be implemented carefully to avoid repeatedly processing the full thought and fact state. A useful system should expose explicit budgets for model updates, source refreshes, and retained bindings.

\paragraph{Privacy and Unobservable Reasoning.}
A latent thought channel may contain sensitive source-derived information even when it is not displayed. Runtime logging, debugging, and model inspection must therefore distinguish operational observability from exposing private chain-of-thought content. Typed metadata and source links can support auditing without requiring full decoding of the thought tensor.

\paragraph{Read-Only Scope.}
The initial CID system focuses on read-only sources such as retrieval, file reads, databases, calculators, clocks, environment state, and streaming feeds. This restriction isolates the central question---whether sustained asynchronous perception can be integrated into a diffusion trajectory---from authorization, rollback, and irreversible action safety. Persistent perception is substantially easier to manage than persistent action. Side-effecting tools require commitment, authorization, idempotency, rollback, and user confirmation mechanisms and are left to future work.

\paragraph{Human Cognition Analogy.}
Sustained perception and repeated cognitive refresh are useful design analogies, not claims that CID models the human brain. CID should be evaluated as a computational architecture on measurable quality, latency, and revision behavior.

\paragraph{Multimodal Extension.}
Although this paper is limited to language and read-only software tools, the separation between externally controlled facts, continuous typed cognition, and display naturally admits visual, audio, and sensor streams. Continuous latent diffusion may provide a shared mechanism for such modalities \citep{guo2026coladlm}, but multimodal training and alignment are future work.

\section{Conclusion}
\label{sec:conclusion}

We introduced Continuous Interaction Diffusion, a model--runtime co-designed architecture that replaces discrete tool-call rounds with persistent asynchronous perception. CID separates externally controlled facts, continuous typed cognition, and user-visible text; represents information needs inside a Typed Cognitive Tensor; and maintains perceptual bindings that can re-project static results or refresh changing sources throughout denoising. CID makes a falsifiable claim: diffusion should let new external information revise existing cognition and display state while useful computation continues. We will test this claim against strong asynchronous autoregressive and explicit-call dLLM baselines.

\clearpage
\bibliographystyle{plainnat}
\bibliography{references}

\clearpage
\appendix
\section{Terminology}
\label{app:terminology}

The following definitions specify how potentially overloaded terms are used throughout this paper.

\begin{description}[leftmargin=0pt,labelindent=0pt,itemsep=2pt,topsep=3pt]
  \item[Autoregressive language model (AR LM).]
  A language model that generates an output by predicting each next token from the input and the already generated prefix. Standard decoding advances left to right and treats earlier emitted tokens as fixed context.

  \item[Diffusion language model (dLLM).]
  A language model that generates through iterative denoising or refinement of a partially unknown or corrupted sequence. Depending on the architecture, several positions may be updated in one step and uncertain earlier positions may remain revisable.

  \item[Denoising step / diffusion update.]
  One iterative update of the diffusion state. A diffusion update can modify multiple positions or cognitive cells and is therefore distinct from generating one token in an AR model.

  \item[Reopening / remasking.]
  Increasing the editability of a previously predicted region so that later computation or newly arrived evidence can revise it. Remasking is one concrete token-level mechanism for reopening.

  \item[Tool call.]
  A discrete executable request to an external source, normally containing a tool or source identity and sufficiently bound arguments such as a query or structured JSON fields.

  \item[Observation.]
  Information returned by an external source or runtime event and made available to the model.

  \item[Information need.]
  A model state indicating that external information would be useful. In CID, such a need may become available to the runtime before it has converged into a complete serialized tool call.

  \item[Perceptual binding.]
  A persistent runtime relation connecting an information need to an external source, its current arguments, refresh policy, cached state, and affected cognitive regions.

  \item[External refresh.]
  Re-executing or re-reading an external source to obtain a potentially newer value. External refresh incurs external I/O or source computation.

  \item[Cognitive refresh / re-projection.]
  Recomputing how an already available external value should influence the current cognitive state, without repeating external I/O.

  \item[Perceptual projection.]
  A context-dependent model representation of an external value, constructed with respect to the current thought and display states and optionally linked to its provenance.

  \item[Fact channel.]
  Externally controlled information that the model may read but may not overwrite. Its contents can still change when the user, environment, runtime, or source updates them.

  \item[Thought channel.]
  The revisable internal cognitive state containing hypotheses, plans, information needs, percept interpretations, uncertainty, constraints, and intermediate conclusions.

  \item[Display channel.]
  The revisable token canvas that progressively converges to the user-visible response.

  \item[Typed Cognitive Tensor (TCT).]
  CID's structured continuous representation of the thought channel. Cognitive cells combine semantic vectors with typed roles, symbolic anchors, source links, uncertainty, local diffusion levels, and lifecycle state.

  \item[Local diffusion level.]
  A per-region value controlling how revisable that region currently is. Higher diffusion permits stronger revision; lower diffusion represents a more stable region.

  \item[Persistent perception.]
  The maintenance of an information source's cognitive influence while the corresponding need remains active. Persistence may reuse a static cached value or refresh a changing source according to policy.

  \item[Static and dynamic sources.]
  A static source is treated as unchanged while a binding is active, allowing repeated cognitive refresh from a cached value. A dynamic source can change over time and may require polling, version checks, subscriptions, or streaming updates.

  \item[Binding lifecycle.]
  The runtime state of a perceptual binding as it moves through stages such as \textsc{candidate}, \textsc{active}, \textsc{waiting}, \textsc{available}, \textsc{refreshing}, and \textsc{retired}.
\end{description}

\section{Reference Runtime Interface}
\label{app:runtime-interface}

A read-only source registered with a CID runtime may expose the following conceptual schema:

\begin{verbatim}
source:
  name: document_search
  description: >-
    retrieve passages from a corpus
  input_schema:
    query: text
    scope: optional identifier
    top_k: optional integer
  properties:
    read_only: true
    cacheable: true
    streamable: true
    cancellable: true
    versioned: true
  default_refresh:
    mode: on_version_change
    max_age_ms: 1000
\end{verbatim}

The runtime should not require that every source support every property. A calculator may be cacheable but not streamable; a clock is dynamic but inexpensive; a file may expose a version identifier; a search endpoint may return incremental results without a stable version.

\section{Binding State Machine}
\label{app:binding-state}

A perceptual binding can use the following lifecycle:
\begin{align}
  \textsc{candidate} &\rightarrow \textsc{active}
  \rightarrow \textsc{waiting} \notag\\
  &\rightarrow \textsc{available}
  \rightarrow \textsc{refreshing} \notag\\
  &\rightarrow \textsc{retired}.
\end{align}
A binding can move from \textsc{available} directly back to \textsc{active} when the value is unchanged and only cognitive re-projection is required. It moves to \textsc{refreshing} when the need remains active and the source freshness policy requires external work. Quiescence is a trajectory state rather than a binding state: active bindings remain intact while model updates pause, and the next observation resumes refinement without reconstructing the interaction context.

A candidate binding should record:
\begin{itemize}[leftmargin=*]
  \item source probability and need confidence;
  \item partially bound selector or arguments;
  \item links to originating cognitive cells;
  \item current cache key and source version;
  \item external and cognitive refresh policies;
  \item target fact, thought, and display regions.
\end{itemize}

\section{Conceptual Execution Traces}
\label{app:traces}

\subsection{Static Copying}

A user requests a response that must reproduce a value from a document exactly. The model forms a persistent need for the value and binds to a file reader. After the file is read once, the runtime caches the exact result. During subsequent denoising steps, the unchanged result is re-projected into the thought tensor, preserving salience while the display is reorganized. No repeated disk I/O is necessary, but the percept remains active until the exact value has been anchored in the display.

\subsection{Dynamic Display Length}

The current realized display length---for example, the inferred EOS position or populated span within the fixed canvas---is an externally computed value that the model may read but cannot rewrite. A binding to the length monitor remains active during generation. At each update, the runtime supplies a refreshed value, which changes the thought tensor's plan and density decisions. The display can compress earlier text because the value conditions both current cognition and already generated output regions.

\subsection{Streaming Retrieval}

A retrieval source first returns titles, then snippets, then complete passages. The binding remains active throughout the stream. Early chunks narrow hypotheses and refine the query; later chunks reopen unsupported conclusions and supply exact evidence. Unrelated display regions remain stable under local diffusion clocks.

\section{Suggested Training Data Construction}
\label{app:data}

A practical initial dataset can be synthesized from ordinary question--answer pairs and documents:
\begin{enumerate}[leftmargin=*]
  \item identify answer spans that depend on external evidence and hide those spans from the model-visible prompt;
  \item create source descriptors, binding targets, and delayed or incremental arrival schedules;
  \item construct pre-arrival states with unresolved cells and post-arrival revisions grounded in the source;
  \item add static-copy cases that preserve one value across many steps and dynamic cases in which the source changes before display convergence.
\end{enumerate}

Teacher-generated textual reasoning may be encoded into initial cognitive cells, but training should randomize cell ordering and event schedules to avoid reducing the thought tensor to a disguised text chain.

\section{Open Technical Questions}
\label{app:questions}

Several design decisions remain open:
\begin{itemize}[leftmargin=*]
  \item which iterative process should update the thought tensor, how many cells it needs, and whether cells should be created or retired dynamically;
  \item whether display and thought diffusion should share a backbone and how source descriptors generalize to unseen tools;
  \item how to estimate support and conflict for local reopening while preserving privacy during debugging;
  \item how to tune refinement-epoch and wall-clock budgets across deployment settings.
\end{itemize}
These questions remain open design variables within the CID framework.

\end{document}